\documentclass[letterpaper]{article} 
\usepackage[submission]{aaai25}  
\usepackage{times}  
\usepackage{helvet}  
\usepackage{courier}  
\usepackage[hyphens]{url}  
\usepackage{graphicx} 
\usepackage{natbib}  
\usepackage{caption} 
\usepackage{amsmath}
\usepackage{svg}
\usepackage{latexsym, graphicx, epsfig, amsmath, amssymb, amsfonts}
\usepackage{amsmath, mathrsfs, mathtools, amsthm}
\usepackage{amssymb}
\usepackage{graphicx}
\usepackage{algorithm}
\usepackage{algorithmic}

\usepackage{newfloat}
\usepackage{listings}
\DeclareMathOperator*{\argmin}{arg\,min}

\DeclareCaptionStyle{ruled}{labelfont=normalfont,labelsep=colon,strut=off} 
\floatstyle{ruled}
\newfloat{listing}{tb}{lst}{}
\floatname{listing}{Listing}
\title{Diffeomorphic Latent Neural Operators for Data-Efficient Learning of Solutions to Partial Differential Equations}
\author {
    Zan Ahmad\textsuperscript{\rm 1} 
    Shiyi Chen\textsuperscript{\rm 2,3}
    Minglang Yin\textsuperscript{\rm 2} 
    Avisha Kumar\textsuperscript{\rm 4}
    Nicolas Charon\textsuperscript{\rm 1,\rm5} \\
    Natalia Trayanova\textsuperscript{\rm 2
    }
    Mauro Maggioni\textsuperscript{\rm 1}
}
\affiliations {
    \textsuperscript{\rm 1}Johns Hopkins University, Department of Applied Mathematics and Statistics\\\textsuperscript{\rm 2}Johns Hopkins University, Department of Biomedical Engineering \\ \textsuperscript{\rm 3}Johns Hopkins University, Department of Mechanical Engineering\\
    \textsuperscript{\rm 4}Johns Hopkins University, Department of Electrical and Computer Engineering\\
    \textsuperscript{\rm 5}University of Houston, Department of Mathematics
}

\usepackage{bibentry}

\begin{document}
\maketitle

\begin{abstract}
A computed approximation of the solution operator to a system of partial differential equations (PDEs) is needed in various areas of science and engineering. Neural operators have been shown to be quite effective at predicting these solution generators after training on high-fidelity ground truth data (e.g. numerical simulations). However, in order to generalize well to unseen spatial domains, neural operators must be trained on an extensive amount of geometrically varying data samples that may not be feasible to acquire or simulate in certain contexts (e.g., patient-specific medical data, large-scale computationally intensive simulations.) We propose that in order to learn a PDE solution operator that can generalize across multiple domains without needing to sample enough data expressive enough for all possible geometries, we can train instead a latent neural operator on just a few ground truth solution fields diffeomorphically mapped from different geometric/spatial domains to a fixed reference configuration. Furthermore, the form of the solutions is dependent on the choice of mapping to and from the reference domain. We emphasize that preserving properties of the differential operator when constructing these mappings can significantly reduce the data requirement for achieving an accurate model due to the regularity of the solution fields that the latent neural operator is training on. We provide motivating numerical experimentation that demonstrates an extreme case of this consideration by exploiting the conformal invariance of the Laplacian.
\end{abstract}

%

\section{Introduction}

Numerical simulation of physical systems governed by partial differential equations (PDEs) are a crucial aspect of many problems in science and engineering. However, traditional numerical methods often demand significant computational resources, particularly for complex systems or high-dimensional parameter spaces. Recent advancements in scientific machine learning include \textit{neural operators}, which can efficiently predict PDE solutions after training on high-fidelity simulation data, as a promising alternative to traditional approaches \cite{kovachki2023neural}. A major challenge in applying neural operators to real-world data lies in the variability of geometric and parametric domains on which these PDEs are defined. Learning solution fields across these varying domains is difficult for typical neural operator frameworks, such as Deep Operator Networks (DeepONet) \cite{li2020fourier} and Fourier Neural Operators (FNO) \cite{lu2021learning} which either require a fixed grid or a regular domain on which the Fourier transform is available, respectively. To address this, some variants of these methods transform numerical solutions from their original domain to a latent space where training can be unified. This process often involves mapping solutions through integral transforms \cite{loeffler2024graph,cao2024laplace,liu2024neural} or spatial transformations \cite{li2023fourier,zhao2024diffeomorphism,yin2024dimon}. Spatial transformations, in particular, standardize the domain by quotienting the geometric variability during training. 

While effective, these approaches require plentiful expressive/geometrically varying training samples to obtain generalizable model. This raises questions about how the choice of transformation influences learning outcomes. Specifically, the properties preserved or distorted by the mapping could significantly affect the neural operator’s ability to learn the PDE efficiently and accurately. The literature has yet to fully address specifically how the nature of the \textit{spatial} transformation of the solution domains and the representation of solutions in the transformed (reference/latent) space impact the learning process.

In this work, we focus on the problem of learning a neural operator on the solution representations after applying an invertible spatial transformation to a reference configuration. Applying this framework on the 2D Laplace equation on doubly-connected domains, we analyze how different types of mappings affect the performance of neural operators and effectively demonstrate that using mappings which preserve mathematical properties of the PDE in the latent space facilitate more accurate and data-efficient learning.

Through numerical experimentation, we compare different mappings of the potential flow solutions to a canonical domain with varying regularity to numerically elucidate the relationship between the choice of mapping, the transformed representation of the PDE solutions, and the strong influence of these factors on the resultant learning efficiency within this framework.  Furthermore, this study aims to provide empirical evidence to motivate the design of mappings (or algorithms that can approximate such mappings) that minimize computational and data requirements for high-fidelity numerical simulations and neural operator training, providing a new avenue for improving data-efficient PDE learning.

\section{Problem Formulation}
\paragraph{Neural Operators for Partial Differential Equations} 
Neural operators are a family of neural network frameworks designed to approximate nonlinear, continuous operators, such as the solution operators of partial differential equations (PDEs). Unlike traditional neural networks, which learn mappings between finite-dimensional vector spaces, neural operators learn mappings between infinite-dimensional function spaces, enabling them to model the relationships between input functions and PDE solutions directly.

Formally, a neural operator \(\mathcal{N}_{\theta}\) is defined as a mapping from an input function space \(\mathcal{V}(\Omega_{\alpha}; \mathbb{R}^{d_v})\) to an output function space \(\mathcal{U}(\Omega_{\alpha}; \mathbb{R}^{d_u})\), where \(\Omega_{\alpha} \subset \mathbb{R}^d\) is a domain parameterized by \(\alpha \in \mathcal{A}\). This mapping can be expressed as:
\[
\mathcal{N}_{\theta}: \mathcal{V}(\Omega_{\alpha}; \mathbb{R}^{d_v}) \to \mathcal{U}(\Omega_{\alpha}; \mathbb{R}^{d_u}),
\]
where \(\theta \in \Theta\) denotes the parameters of the neural operator. The spaces \(\mathcal{V}(\Omega_{\alpha}; \mathbb{R}^{d_v})\) and \(\mathcal{U}(\Omega_{\alpha}; \mathbb{R}^{d_u})\) are subspaces of function spaces such as Sobolev spaces \(H^s(\Omega_{\alpha})\) or spaces of continuous functions \(C(\Omega_{\alpha})\). Here:
\begin{itemize}
    \item \(\mathcal{V}(\Omega_{\alpha}; \mathbb{R}^{d_v})\) represents the space of input functions, such as coefficients, forcing terms, or boundary conditions of the PDE.
    \item \(\mathcal{U}(\Omega_{\alpha}; \mathbb{R}^{d_u})\) corresponds to the space of PDE solutions.
\end{itemize}

For an input \((v_1, \ldots, v_m) \in \mathcal{V}(\Omega_{\alpha}; \mathbb{R}^{d_v})\), the true solution operator \(\mathcal{G}\) maps these inputs to the PDE solution \(u \in \mathcal{U}(\Omega_{\alpha}; \mathbb{R}^{d_u})\), satisfying:
\[
u = \mathcal{G}(v_1, \ldots, v_m).
\]

The neural operator \(\mathcal{N}_{\theta}\) approximates the solution operator \(\mathcal{G}\) by learning from a set of input-output pairs:
\[
\{(v_{1,i}, \ldots, v_{m,i}, u_i)\}_{i=1}^N, \quad \text{where } u_i = \mathcal{G}(v_{1,i}, \ldots, v_{m,i}).
\]
Once trained, \(\mathcal{N}_{\theta}\) serves as a surrogate for \(\mathcal{G}\), enabling efficient and accurate predictions of \(u\) for new inputs \((v_1, \ldots, v_m)\) without the computational cost of solving the PDE numerically. By working directly with function spaces, neural operators offer the flexibility and efficiency needed for tasks involving parametric PDE simulations and high-dimensional domains.

\subsection{Diffeomorphic Mapping Operator Learning}
This section outlines the approach we adapted from \cite{yin2024dimon} for learning operators on a family of domains for which between each pair there exists a diffeomorphism. The procedure involves mapping solutions from a set of geometrically diverse domains \(\{\Omega_{\alpha}\}_{\alpha \in \mathcal{A}}\) to a fixed reference domain \(\Omega_0\). These mappings, defined by smooth embeddings \(\varphi_{\alpha}: \Omega_{\alpha} \to \Omega_0\), allow solutions of partial differential equations (PDEs) on different domains to be consistently represented within a common space.

Let \(\mathcal{A}\) denote a compact subset of \(\mathbb{R}^p\), parametrizing the domain variations. For each \(\alpha \in \mathcal{A}\), the embedding \(\varphi_{\alpha}\) is a \(C^2\) diffeomorphism from the reference domain \(\Omega_0\) to the target domain \(\Omega_{\alpha}\), and the mapping \(\mathcal{A} \to C^2(\Omega_0, \mathbb{R}^d)\), given by \(\alpha \mapsto \varphi_{\alpha}\), is continuous in the standard \(C^2\)-norm.

The solution operator on each domain \(\Omega_{\alpha}\) is denoted by \[\mathcal{G}_{\alpha}: \mathcal{V}^{\alpha}_1(\Omega_{\alpha}) \times \cdots \times \mathcal{V}^{\alpha}_m(\Omega_{\alpha}) \to \mathcal{U}^{\alpha}(\Omega_{\alpha})\] mapping inputs defined on \(\Omega_{\alpha}\) to the PDE solution \(u^{\alpha}(\cdot; \mathbf{v}^{\alpha})\). Here, \(\mathbf{v}^{\alpha} = (v_1^{\alpha}, \ldots, v_m^{\alpha})\) represents the input functions.
\begin{figure}
    \centering
    \includegraphics[width=1\linewidth]{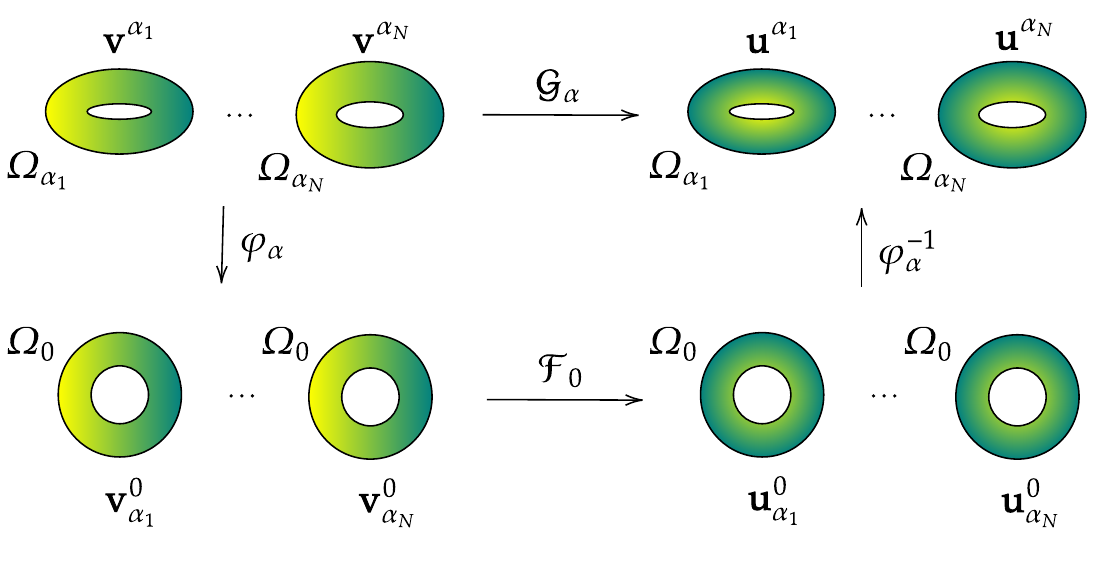}
    \caption{A schematic outlining the construction of the latent operator $\mathcal{F}_0$ within the diffeomorphic mapping operator learning framework.}
    \label{fig:dimon}
\end{figure}

This family of operators \(\{\mathcal{G}_{\alpha}\}_{\alpha \in \mathcal{A}}\) can be reformulated in terms of a latent operator \(\mathcal{F}_0\) defined on the reference domain \(\Omega_0\), as:
\begin{align*}
u^\alpha(\mathbf{x}) &= \mathcal{G}_{\alpha}(v_1^{\alpha}, \ldots, v_m^{\alpha}) \\
 &= \mathcal{F}_0\left(\alpha, v_1^{\alpha} \circ \varphi_{\alpha}, \ldots, v_m^{\alpha} \circ \varphi_{\alpha}\right) \circ \varphi_{\alpha}^{-1}.
\end{align*}

Defining the pullback of the inputs as \(v_{\alpha,k}^0 = v_k^{\alpha} \circ \varphi_{\alpha}\) and the latent solution as \(u^0_{\alpha} = \mathcal{F}_0(\alpha, v_{\alpha,1}^0, \ldots, v_{\alpha,m}^0)\), the solution on \(\Omega_{\alpha}\) can be equivalently written as:
\[
u^{\alpha}(\cdot; \mathbf{v}^\alpha) = u^0_{\alpha} \circ \varphi_{\alpha}^{-1}.
\]

Here, the operator \(\mathcal{F}_0\) captures the relationship between the shape parameter \(\alpha\), the input functions transported to \(\Omega_0\), and the corresponding solution on the reference domain. This formulation effectively maps solutions from \(\Omega_{\alpha}\) into a fixed domain \(\Omega_0\), enabling the neural operator to learn in a geometry-independent space. Of course, the representation of the solution (for a fixed $\alpha \in \mathcal{A}$) in the latent space where a neural operator learns $\mathcal{F}_0$ is dependent on the choice of map and (in most cases) the specific geometries 
$\Omega_{\alpha}$ and $\Omega_{0}$. In principle, $\mathcal{F}_0$ may be approximated by any neural operator method for the chosen $\Omega_0$. 

In this work, we use a modified Deep Operator Network (DeepONet) architecture similar to MIONet \cite{jin2022mionet} to approximate the solution operator of the form:
\[
\mathcal{F}_0: \mathcal{A} \times \mathcal{V}_1(\Omega_0) \times \cdots \times \mathcal{V}_m(\Omega_0) \to \mathcal{U}(\Omega_0),
\]
where \(\mathcal{V}_k(\Omega_0)\) and \(\mathcal{U}(\Omega_0)\) are function spaces defined on a fixed grid \(\Omega_0\), and \(\mathcal{A}\) represents the parameter space.

Our architecture consists of two key input encodings:
\begin{itemize}
    \item A geometry encoder \(\mathbf{g}_{\text{geo}}(\alpha): \mathcal{A} \to \mathbb{R}^q\), which maps the shape parameter \(\alpha\) to a low-dimensional representation. This encoding can be computed using principal component analysis (PCA) on the displacement vectors between the target domain \(\Omega_{\alpha}\) and the reference domain \(\Omega_0\), or by computing a representation of the embedding directly from the flow of the map \(\varphi_{\alpha}\).
    \item A physical condition encoder \(\mathbf{g}_{\text{phys}}(\mathbf{v}_{\alpha}^0): \mathcal{V}_1(\Omega_0) \times \cdots \times \mathcal{V}_m(\Omega_0) \to \mathbb{R}^r\), which encodes the initial and boundary conditions transported to \(\Omega_0\).
\end{itemize}

These encodings are concatenated and used as inputs to a neural operator. For spatial dependence, we use a separate network \(\mathbf{h}(\mathbf{x}): \Omega_0 \to \mathbb{R}^s\), which maps spatial coordinates to a feature representation. The final approximation of the solution is given by:
\[
u^0_{\alpha}(\mathbf{x}) \approx \hat{\mathcal{F}}_\theta \left( \sum_{i=1}^n \mathbf{w}_i\big(\mathbf{g}_{\text{geo}}(\alpha), \mathbf{g}_{\text{phys}}(\mathbf{v}_{\alpha}^0)\big) \cdot \mathbf{h}_i(\mathbf{x}) \right),
\]
where:
\begin{itemize}
    \item \(\mathbf{g}_{\text{geo}}(\alpha)\) encodes the geometric information for the shape parameter \(\alpha\) and \(\theta\) denotes the network parameters,
    \item \(\mathbf{g}_{\text{phys}}(\mathbf{v}_{\alpha}^0)\) encodes the transported initial and boundary conditions \(\mathbf{v}_{\alpha}^0\),
    \item \(\mathbf{h}_i(\mathbf{x})\) evaluates the spatial basis functions at \(\mathbf{x} \in \Omega_0\),
    \item \(\mathbf{w}_i\big(\mathbf{g}_{\text{geo}}(\alpha), \mathbf{g}_{\text{phys}}(\mathbf{v}_{\alpha}^0)\big)\) computes the coefficients by combining the outputs of the geometry and physical condition encoders.
\end{itemize}

This architecture is well-suited to our framework since all solutions and inputs are mapped to the fixed reference domain \(\Omega_0\) and fixed grid at that, enabling the network to learn the latent operator \(\mathcal{F}_0\) efficiently. By incorporating shape information and input conditions through separate encoders, the network captures the dependence of the solution operator on both geometry and PDE parameters.

\subsection{Latent Solution Representations}

The choice of spatial transformation \(\varphi_{\alpha}: \Omega_{\alpha} \to \Omega_0\) plays a pivotal role in the representation of the latent solution \(u^0_{\alpha}\) on the reference domain \(\Omega_0\) and the form of the \textit{diffeomorphic latent operator} \(\mathcal{F}_0\). While it is not always possible to find a map that perfectly preserves the differential operator \(\mathcal{L}_{\alpha}\) of a PDE, one can aim to construct maps that minimize deviations from this ideal. Such maps improve the regularity of the latent solution representation and enable a more accurate approximation of the true solution operator. They also decrease the sensitivity of the mapped solution fields to changes in the shape parameter $\alpha$ of the original domain $\Omega_{\alpha}$ and lower the required dimensionality of the shape encoding ``geometry branch" in the DeepONet for efficiently approximating $\mathcal{F}_0$.

\paragraph{Optimal Maps for PDE Preservation} 
Consider the PDE on \(\Omega_{\alpha}\) given by:
\[
\mathcal{L}_{\alpha}[u^{\alpha}] = v_{\alpha}, \quad u^{\alpha} \in \mathcal{U}(\Omega_{\alpha}),
\]
where \(\mathcal{L}_{\alpha}\) is a differential operator and \(v_{\alpha} \in \mathcal{V}_k(\Omega_{\alpha})\) represents the inputs. Under the transformation \(\varphi_{\alpha}\), this PDE is pulled back to the reference domain \(\Omega_0\), resulting in:
\[
\tilde{\mathcal{L}}_{\alpha}[u^0_{\alpha}] = v_{\alpha} \circ \varphi_{\alpha}, \quad u^0_{\alpha} \in \mathcal{U}(\Omega_0),
\]
where \(\tilde{\mathcal{L}}_{\alpha}\) is the transformed operator. The latent operator \(\mathcal{F}_0\) provides the strongest approximation to the solution operator \(\mathcal{G}_{\alpha}\) when the map \(\varphi_{\alpha}\) minimizes deviations between \(\tilde{\mathcal{L}}_{\alpha}\) and \(\mathcal{L}_{\alpha}\).

To identify such a map, one can solve a minimization problem that balances the preservation of the PDE operator and the fidelity of the geometric mapping:
\[
\varphi_{\alpha}^* = \argmin_{\varphi_{\alpha}} \mathcal{E}[\varphi_{\alpha}]\]\[\\ \quad \mathcal{E}[\varphi_{\alpha}] = \int_{\Omega_0} \|\tilde{\mathcal{L}}_{\alpha}[u^0_{\alpha}] - (\mathcal{L}_{\alpha}[u^{\alpha}] \circ \varphi_{\alpha})\|^2 \, dx + \lambda \, \mathcal{D}[\varphi_{\alpha}],
\]
where \(\mathcal{D}[\varphi_{\alpha}]\) is a regularization term that enforces the geometric alignment of \(\varphi_{\alpha}\) with \(\Omega_{\alpha}\), and \(\lambda\) is a weighting parameter. This optimization problem seeks to find a near-optimal map \(\varphi_{\alpha}^*\) that preserves the differential operator as closely as possible while accurately mapping the reference domain \(\Omega_0\) to the target domain \(\Omega_{\alpha}\).

\paragraph{Impact of Reference Domain Choice}  
The choice of reference domain \(\Omega_0\) also influences the latent solution representation \(u^0_{\alpha}\). A well-chosen reference domain can simplify the geometry of the target domains \(\{\Omega_{\alpha}\}_{\alpha \in \mathcal{A}}\) under the transformations \(\varphi_{\alpha}\), improving the regularity of the latent solution \(u^0_{\alpha}\). Conversely, a poorly chosen reference domain may introduce distortions in the transformed operator \(\tilde{\mathcal{L}}_{\alpha}\), complicating the representation of \(\mathcal{F}_0\).

The selection of \(\Omega_0\) should therefore account for the geometric and functional characteristics of the family of domains \(\{\Omega_{\alpha}\}\) and the PDE operator \(\mathcal{L}_{\alpha}\). For example, in the case of the Laplace equation, using a reference domain conformally equivalent to the target domains ensures that the latent solutions \(u^0_{\alpha}\) remain well-behaved, as the Laplace operator is invariant under conformal transformations.

\paragraph{Regularity of Latent Solutions}  
By constructing maps that minimize deviations from PDE invariance and selecting an appropriate reference domain, one can improve the regularity of the latent solutions \(u^0_{\alpha}\). Regularity here refers to the smoothness and stability of the latent solutions with respect to variations in the input parameters \(\alpha\). Improved regularity reduces the complexity of the neural operator \(\mathcal{F}_0\), enabling more efficient training and better generalization to unseen parameter configurations.

This work sheds light on the relationship between the choice of \(\varphi_{\alpha}\), the reference domain \(\Omega_0\), and the latent operator \(\mathcal{F}_0\), emphasizing the importance of designing spatial transformations that align with the underlying physical and geometric properties of the PDE.
\section{Numerical Experiment}
In this section, we describe the example we use to illustrate the main ideas of the neural operator approach described in the previous section. Specifically, we construct three different mappings $\varphi_\alpha:\Omega_\alpha \rightarrow \Omega_0$ with different degrees of regularity on the mapped solutions and compare how well and how efficiently each of them allowed the latent neural operator to learn $\mathcal{F}_0$ and, via composition with $\varphi_\alpha^{-   1}$, $\mathcal{G}_\alpha$
\subsection{Laplace Equation}
We use the 2D Laplace equation as a test case for our neural operator framework. 2D Laplace equation emerges in many fields of science and engineering. For example, in fluid dynamics, it can represent the solution of potential flow where both vorticity and viscosity are considered negligible. This simplification allows us to model the velocity field as the gradient of a scalar potential field, $u(x, y)$. Under the assumptions of incompressibility, the governing equation for the potential field reduces to the Laplace equation:
\begin{align}
    \nabla^2 u &= 0 \quad \text{ on } \Omega_{\alpha}\\
    \mathbf{n} \cdot \nabla u &= 0 \quad \text{ on } \partial \Omega^I_{\alpha}\\
    \mathbf{n} \cdot \nabla u &= b_{\alpha}(s) \quad \text{ on } \partial \Omega^o_{\alpha}
    \label{bc}
\end{align}

where the boundary $\partial\Omega_\alpha$ is the union of the inner and outer boundaries:
\begin{equation}
    \partial\Omega_\alpha =  \partial\Omega_\alpha^I \cup\partial\Omega_\alpha^o
\end{equation}

and the $b_{\alpha}(s)$ in Equation~\ref{bc} represents the Neumann Boundary Condition on the outer side of the disk. 

$\mathbf{n}$ is the normal vector perpendicular to $\partial \Omega_{\alpha}$ for both boundaries. 





\subsubsection{Compatibility Condition}
For the Laplace equation to possess a valid solution, the specified boundary conditions must satisfy a compatibility condition. The condition is represented as:
\begin{equation}
    \int_{\Omega_{\alpha}} \nabla^2 u\, dA  = \int_{\partial\Omega^o_{\alpha}} \mathbf{n} \cdot \nabla u\, ds  =  \int_{\partial\Omega^o_{\alpha}} b_{\alpha}(s)\, ds = 0
\end{equation}

where the second equality is derived from the Divergence theorem. 

\subsection{Finite Elements for Laplace Equation}
The Laplace equation $\nabla^2 u  = 0$ can be solved in weakly by:
\begin{equation}
	\int_{\Omega_{\alpha}} \nabla^2u \cdot \tilde{u}\,  dx = 0
\end{equation}
We let $u \in U$ be the trial function and $\tilde{u} \in U$ be the test function, where $U$ is the function space defined on the target domain:
\begin{equation}
	U = \mathcal{U}(\Omega_{\alpha}; \mathbb{R}^2)
\end{equation}
Applying the divergence theorem:
\begin{equation}
	\begin{aligned}
		0&=\int_{\Omega_{\alpha}} \nabla^2u \cdot \tilde{u}\, dx = \int_{\Omega_{\alpha}} \nabla \cdot (\nabla u) \cdot \tilde{u}  dx \\
		&= \int_{\partial\Omega_{\alpha}} \langle\nabla u, \mathbf{n}\rangle  \tilde{u}  ds - \int_{\Omega_{\alpha}} \langle\nabla u, \nabla \tilde{u}\rangle  dx
	\end{aligned}
\end{equation}
The variational form of the Laplace equation reads: find $u \in U$ such that
\begin{equation}
	a(u, v) = L(\tilde{u}) \quad \forall \tilde{u} \in U
\end{equation}
where the bilinear form $a(u,v)$ and linear form $L(v)$ are defined as:
\begin{equation}
	\begin{aligned}
		a(u, \tilde{u}) &= \int_{\Omega_{\alpha}} \langle\nabla u, \nabla \tilde{u}\rangle  dx \ =\\
		L(v) &= \int_{\partial\Omega_{\alpha}^o} b \cdot v  ds - \int_{\partial\Omega_{\alpha}^I} 0 \cdot \tilde{u}  ds
	\end{aligned}
\end{equation}
\subsection{Domain Generation}
The foundation of our shape generation method lies in a modified Joukowski-type transformation \cite{karpfinger2022conformal}. The basic mapping function takes the form:
\begin{equation}
	J(z) = z + \frac{a^2}{z}
\end{equation}
where $a$ is a parameter satisfying $a > 1$. This transformation is a variation of the classical Joukowski mapping, which is historically significant in aerodynamics for generating airfoil profiles. In our implementation, we introduce randomness to the parameter $a$:

\begin{equation}
	a = 1.1 + 0.2\xi_0
\end{equation}

where $\xi_0$ is a uniform random variable on $[0,1]$. 

The base shape is generated by applying this transformation to points on the unit circle with center $ x_c + i y_c  $. The center are generate randomly by:

\begin{equation}
	x_c = 0.3(\xi_1 - 0.5)
\end{equation}
\begin{equation}
	y_c = 0.3(\xi_2 - 0.5)
\end{equation}

where $\xi_1$ and $\xi_2$ are independent uniform random variables on $[0,1]$. The transformed points are thus:

\begin{equation}
	z_1 = (e^{i\theta} + x_c + iy_c)
\end{equation}

\begin{equation}
	\partial\Omega^I_{\alpha} = J(z_1) = z_1 + \frac{a^2}{z_1}
\end{equation}
In our case, the domain shape parameter $\alpha$ has three degrees of freedom, they are $a, \xi_1, \xi_2. $
\subsection{2D Conformal Mapping}
Conformal mapping is a transformation that preserves angles and the local shapes of structures. In the context of our application, conformal mapping has a particularly useful property: if a function \( g \) is harmonic, i.e., \( \nabla^2 g = 0 \), and \( \varphi \) is a conformal mapping on the 2D plane, then the transformed function \( \varphi \circ g \) is also harmonic \cite{parker1987invariants}. This property, known as the \textit{conformal invariance of the Laplacian}, makes conformal mappings especially valuable when working with harmonic PDEs such as the Laplace equation.

\begin{figure}
    \centering
    \includegraphics[width=1.0\linewidth]{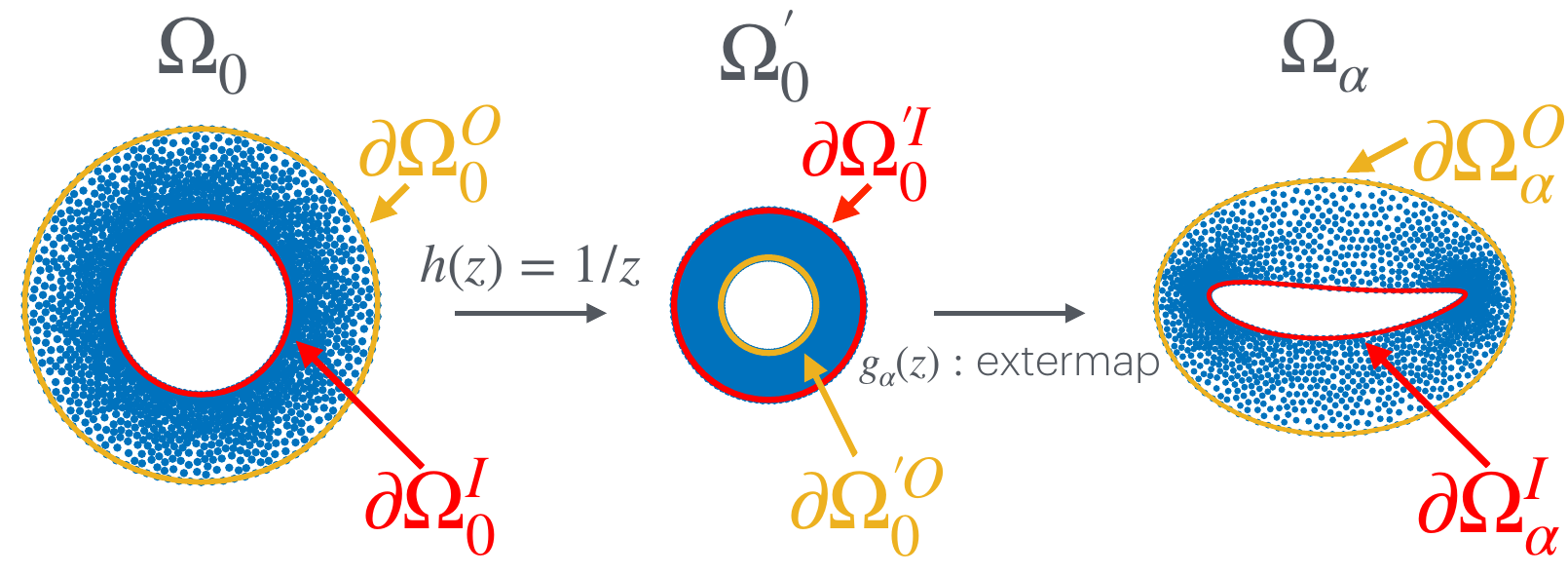}
    \caption{Schematics for $\varphi^{-1}$ computation via the \texttt{extermap} function in sc-toolbox.}
    \label{scmap}
\end{figure}

Our objective is to find a conformal mapping from the target domains \(\{\Omega_{\alpha}\}\) to a fixed reference domain \(\Omega_0\). Conformal mappings are generally difficult to compute analytically, especially for arbitrary shapes. To overcome this, we compute these mappings numerically using the MATLAB Schwarz–Christoffel Toolbox \cite{driscoll1996algorithm}, developed by Toby Driscoll. The Schwarz–Christoffel (SC) mapping is a powerful method for finding conformal maps from the unit disk or upper half-plane to polygonal domains (Figure \ref{scmap}). 
 
The Schwarz–Christoffel mapping is expressed as \cite{driscoll2002schwarz}:
\begin{equation}
    \varphi(z) = C + \int^z \prod_{k=1}^n (w - z_k)^{\omega_k - 1} \, dw,
\end{equation}
where:
\begin{itemize}
    \item \( z_k \) are the pre-images of the polygon's vertices on the unit disk or upper half-plane,
    \item \( \omega_k \) are the interior angles at the vertices, scaled relative to \(\pi\),
    \item \( C \) is an additive constant of integration.
\end{itemize}

This integral describes a holomorphic function that maps the source domain (e.g., the unit disk) to a polygonal target domain, preserving angles at the vertices. For doubly connected domains, SC mappings require extensions to account for inner and outer boundaries, which are handled by numerical tools such as the Schwarz–Christoffel Toolbox.

Let $\partial\Omega^I$ denote the inner boundary and $\partial \Omega^o$ denotes the outer boundary. Since the radius in our reference domain $r$ is fixed, we must generate conformally equivalent shapes in our numerical experiment. After generating $\partial\Omega^I$ using the Jowkowsky approach, We utilize \texttt{extermap} in sc-toolbox to approximate a mapping from a unit disk to the exterior of $\partial\Omega^I.$ Figure~\ref{scmap} shows the boundaries of original and mapped domains. From $\Omega_0$ to $\Omega_\alpha,$ we first invert the inner and outer radius by $h(z) = \frac{1}{z}$ Since $\partial\Omega^I_0$ is a unit circle, it is invariant under $h(z),$ and $\partial\Omega^I_0 = \partial\Omega^{'I}_0.$ Based on the shape of the target inner boundary $\partial \Omega^I_\alpha,$ we use \texttt{extermap} to calculate a Schwarz–Christoffel mapping $g_{\alpha}$ that maps a unit circle to the exterior of $\partial \Omega^I_\alpha.$ Then the outer boundary $\partial \Omega^O_0$ is mapped passively by the composition $\varphi^{-1}_{\alpha} = g_{\alpha} \circ h: $

\begin{equation}
    \partial \Omega^o_\alpha = \varphi^{-1}_{\alpha}(\partial \Omega^o_0) = g_{\alpha} \circ h(\partial \Omega^o_0) 
\end{equation}

We leave the outer domain free to deform under $f$ make sure that the domains we generated are conformally equivalent to each other.

By leveraging conformal mappings, we transform solutions \( u^{\alpha} \in \mathcal{U}(\Omega_{\alpha}) \) defined on doubly connected target domains to a fixed reference annulus \(\Omega_0 = \mathbb{A}_r\). The mappings are computed numerically and used to standardize the input and solution spaces. Specifically, for a given target domain \(\Omega_{\alpha}\), we compute the pullback of the input and solution functions via the conformal mapping \( \varphi_\alpha \):
\begin{equation}
v_{\alpha,k}^0 = v_k^{\alpha} \circ \varphi^{-1}_{\alpha}, \quad u^0_{\alpha} = u^{\alpha} \circ \varphi^{-1}_{\alpha}.
\end{equation}

This ensures that the latent neural operator for approximating \(\mathcal{F}_0\) learns solutions in a geometry-independent space while preserving the harmonic structure of the transformed solutions due to the conformal invariance of the Laplacian. In theory, since the geometry is completely factored out, the only relationship we need to learn is between the boundary condition $f(s)$ and the solution representation in $\mathcal{F}_0$. 

\subsection{Large Deformation Diffeomorphic Metric Mapping (LDDMM)}

Large Deformation Diffeomorphic Metric Mapping (LDDMM) is a framework for constructing smooth, invertible transformations (diffeomorphisms) between shapes or domains \cite{beg2005computing}. It models these transformations as flows induced by time-dependent vector fields and introduces a geometric structure to the space of diffeomorphisms. In our work, LDDMM provides the transformations \(\varphi_{\alpha}: \Omega_{\alpha} \to \Omega_0\) needed to map the reference domain \(\Omega_0\) to a family of target domains \(\{\Omega_{\alpha}\}_{\alpha \in \mathcal{A}}\). These transformations standardize geometric variations, enabling a neural operator to work in a unified latent space.

LDDMM constructs diffeomorphic transformations as flows of time-dependent velocity fields, where the velocity field is represented as the time derivative of the flow map, \(\dot{\gamma}_t \in \Gamma\). Here, \(\Gamma\) is a Hilbert space of sufficiently smooth vector fields, such as a Sobolev or reproducing kernel Hilbert space (RKHS). The associated flow map \(\gamma_t\) satisfies the following ordinary differential equation (ODE):
\begin{equation}
\label{eq:flow}
\frac{d}{dt} \gamma_t(x) = \dot{\gamma}_t(\gamma_t(x)), \quad \gamma_0(x) = x, \quad t \in [0,1],
\end{equation}
where \(\gamma_t: \mathbb{R}^d \to \mathbb{R}^d\) is a family of diffeomorphisms indexed by time \(t\). The diffeomorphism at terminal time \(t = 1\), \(\gamma_1\), provides the desired map between \(\Omega_0\) and \(\Omega_{\alpha}\).

The deformation energy of the flow \(\dot{\gamma}\) is given by:
\begin{equation}
\label{eq:lddmm_energy}
E[\dot{\gamma}] = \int_0^1 \|\dot{\gamma}_t\|_{\Gamma}^2 \, dt,
\end{equation}
where \(\|\dot{\gamma}_t\|_{\Gamma}\) is the norm of \(\dot{\gamma}_t\) in \(\Gamma\). This energy quantifies the cost of the deformation. Given two domains \(\Omega_0\) and \(\Omega_{\alpha}\), the optimal transformation is computed by solving:
\begin{equation}
\label{eq:LDDMM_matching}
\inf_{\dot{\gamma} \in L^2([0,1], \Gamma)} \left\{ \int_0^1 \|\dot{\gamma}_t\|_{\Gamma}^2 \, dt \ \middle| \ \gamma_1(\Omega_0) = \Omega_{\alpha} \right\}.
\end{equation}

In practice, the hard constraint \(\gamma_1(\Omega_0) = \Omega_{\alpha}\) is often relaxed by introducing a similarity measure \(\mathcal{D}[\varphi_{\alpha}]\) between \(\gamma_1(\Omega_0)\) and \(\Omega_{\alpha}\), resulting in the minimization:
\begin{equation}
\inf_{\dot{\gamma} \in L^2([0,1], \Gamma)} \left\{ \int_0^1 \|\dot{\gamma}_t\|_{\Gamma}^2 \, dt + \lambda \, \mathcal{D}[\varphi_{\alpha}] \right\},
\end{equation}
where \(\lambda > 0\) is a weighting parameter that balances deformation energy and geometric alignment.

The optimization problem can be approached using principles from optimal control. Specifically, the Pontryagin maximum principle provides conditions that the optimal \(\dot{\gamma}\) must satisfy \cite{younes2010shapes}. These conditions show that \(\dot{\gamma}\) can be parameterized by an initial momentum variable. For discrete shapes, where the domains are represented as finite sets of points or vertices, the initial momentum is distributed over these points. This reduces the problem to a finite-dimensional optimization problem over the initial momentum, often solved using a shooting approach.

For the applications presented in this paper, we employ the \texttt{FshapesTk} MATLAB library for diffeomorphic registration, which implements LDDMM for both continuous and discrete domains. (The software can be found at \url{https://plmlab.math.cnrs.fr/benjamin.charlier/fshapesTk}.)

In this work, LDDMM provides the diffeomorphic transformations \(\varphi_{\alpha} = \gamma_1\) required to map the reference domain \(\Omega_0\) to each target domain \(\Omega_{\alpha}\). These maps standardize the domains, enabling consistent representation of input functions and solutions in the reference domain:
\begin{align*}
v_{\alpha,k}^0 &= v_k^{\alpha} \circ \varphi_{\alpha}, \quad k = 1, \ldots, m, \\
u^0_{\alpha} &= u^{\alpha} \circ \varphi_{\alpha}.
\end{align*}
This transformation ensures that the neural operator \(\mathcal{F}_0\) learns a unified mapping that is independent of geometric variability.

The diffeomorphisms generated by LDDMM are guaranteed to be smooth and invertible, preserving the topological properties of the domains. However, these maps do not necessarily preserve the differential operator \(\mathcal{L}_{\alpha}\) of the PDE, potentially introducing distortions in the transformed solutions \(u^0_{\alpha}\). Furthermore, since LDDMM maps are not conformal, they do not preserve angles, which is crucial for maintaining the invariance of the Laplacian under transformations.

\subsection{Discrete Optimal Transport Mapping}
As an example of a non-diffeomorphic mapping, we consider the use of discrete optimal transport (OT) to map solutions from varying domains to a fixed reference domain \cite{peyre2019computational}. This approach computes a pointwise bijection between point clouds representing the domains, ensuring uniform mass transport under a specified cost function. Unlike diffeomorphic mappings, discrete optimal transport does not guarantee smoothness, making it a useful benchmark method for training our diffeomorphic latent neural operators.

Let \( \Omega_0 \) and \( \Omega_{\alpha} \) represent the reference and target domains, respectively, and let their boundary points or interior points be discretized into two point clouds:
\[
X = \{x_1, x_2, \dots, x_N\} \subset \mathbb{R}^d, \]\[Y = \{y_1, y_2, \dots, y_N\} \subset \mathbb{R}^d,
\]
where \( N \) is the number of points in each cloud. Assume both point clouds have uniform measures:
\[
\mu_X = \frac{1}{N} \sum_{i=1}^N \delta_{x_i}, \quad \mu_Y = \frac{1}{N} \sum_{i=1}^N \delta_{y_i},
\]
where \(\delta_{x_i}\) is the Dirac delta measure at \(x_i\).

The discrete optimal transport problem seeks a bijective mapping \(T: X \to Y\) that minimizes the total transport cost. This can be formulated as:
\[
\min_{\pi \in \Pi} \sum_{i=1}^N \sum_{j=1}^N c(x_i, y_j) \pi_{ij},
\]
where:
\begin{itemize}
    \item \( \pi = (\pi_{ij}) \in \mathbb{R}^{N \times N} \) is the transport plan, where \( \pi_{ij} \) represents the fraction of mass transported from \(x_i\) to \(y_j\),
    \item \( c(x_i, y_j) = \|x_i - y_j\|^2 \) is the transport cost between points \(x_i\) and \(y_j\),
    \item \( \Pi \) is the set of admissible transport plans:
\end{itemize}
 \[
     \left\{ \pi \in \mathbb{R}^{N \times N} \ \middle| \ \sum_{j=1}^N \pi_{ij} = \frac{1}{N}, \ \sum_{i=1}^N \pi_{ij} = \frac{1}{N}, \ \pi_{ij} \geq 0 \ \forall i, j \right\}
    \]
If the cost function \(c(x_i, y_j)\) is strictly convex (e.g., quadratic cost), the optimal transport plan \(\pi^*\) is a permutation matrix. This corresponds to a bijective mapping \(T: X \to Y\) defined as:
\[
T(x_i) = y_j \quad \text{if} \quad \pi_{ij}^* = 1.
\]

Since the optimal transport plan \(\pi^*\) is a permutation matrix, the resulting mapping \(T\) is invertible. The inverse map \(T^{-1}: Y \to X\) satisfies:
\[
T^{-1}(y_j) = x_i \quad \text{if} \quad \pi_{ij}^* = 1.
\]

In this work, we compute the optimal transport map using the Hungarian algorithm, which efficiently finds the minimum-cost assignment in \(\mathcal{O}(N^3)\) time \cite{kuhn1955hungarian}. Given the uniform measures on the point clouds, the algorithm outputs a bijection \(T\) that minimizes the transport cost.

We apply the computed transport map \(T\) to align solutions from \(\Omega_{\alpha}\) with the reference domain \(\Omega_0\). Specifically, given an input function \(v^{\alpha} \in X_k(\Omega_{\alpha})\) and solution \(u^{\alpha} \in Y(\Omega_{\alpha})\), their pullbacks to the reference domain are:
\[
v_{\alpha,k}^0 = v_k^{\alpha} \circ T^{-1}, \quad u^0_{\alpha} = u^{\alpha} \circ T^{-1}.
\]
These transformed functions are used to train the latent operator \(\mathcal{F}_0\) on the reference domain.

The discrete optimal transport mapping is computationally efficient and guarantees a bijective alignment of point clouds. However, as it lacks smoothness, the resulting transformations may distort the spatial structure of solutions, introducing artifacts that affect learning. This makes it a non-diffeomorphic yet useful benchmark for comparison with smooth transformations, such as those generated by LDDMM or conformal mappings.
\section{Results and Discussion}
\begin{figure*}[h]
    \centering
    \includegraphics[width=1\linewidth, trim=0 100 0 150, clip]{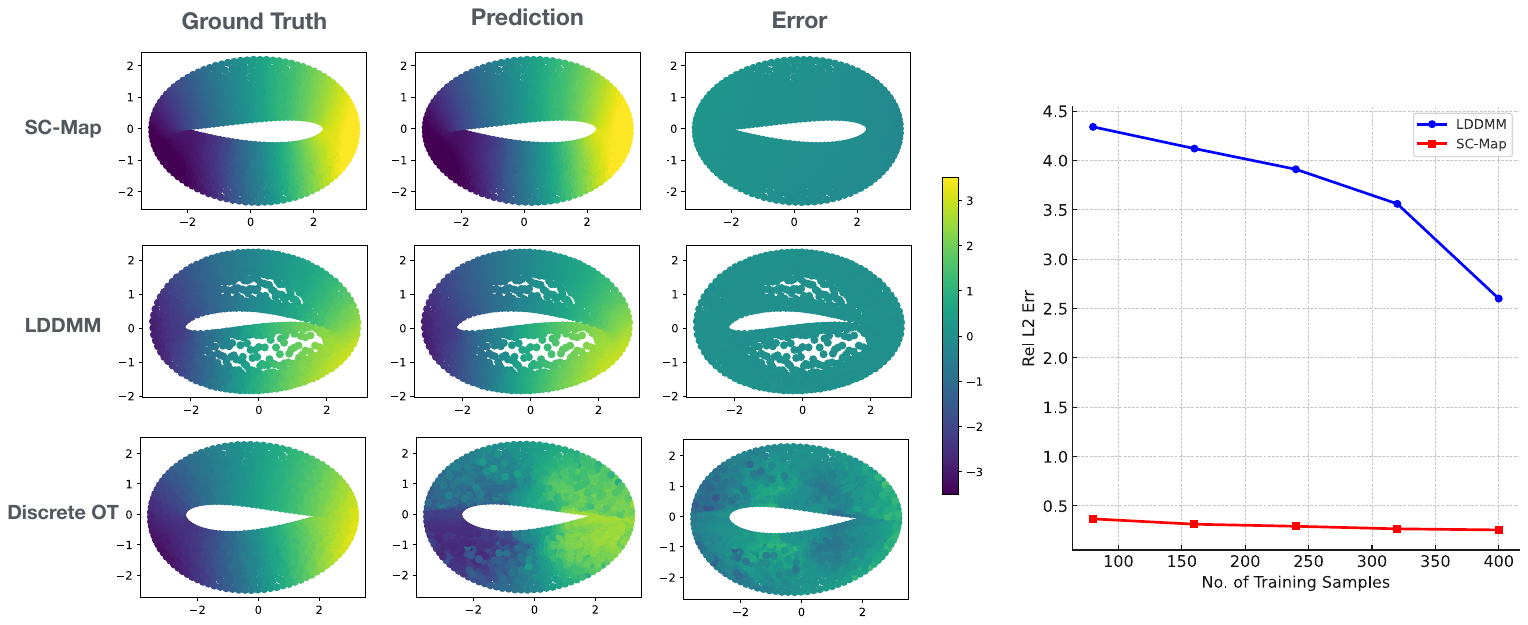}
    \vspace{-35pt} 
    \caption{Comparison of mapping approaches (SC-Map, LDDMM, and Discrete OT) for the Laplace solutions on doubly connected domains. The first three columns illustrate the ground truth solution, predicted solution, and the error (ground truth - prediction) for each mapping approach. The rightmost panel shows the relative \(L_2\) error as a function of the number of training samples, highlighting the superior performance and efficiency of SC-Map compared to LDDMM.}
    \label{fig:results-mega}
\end{figure*}

\begin{figure}[h]
    \centering
    \includegraphics[width=0.81\linewidth]{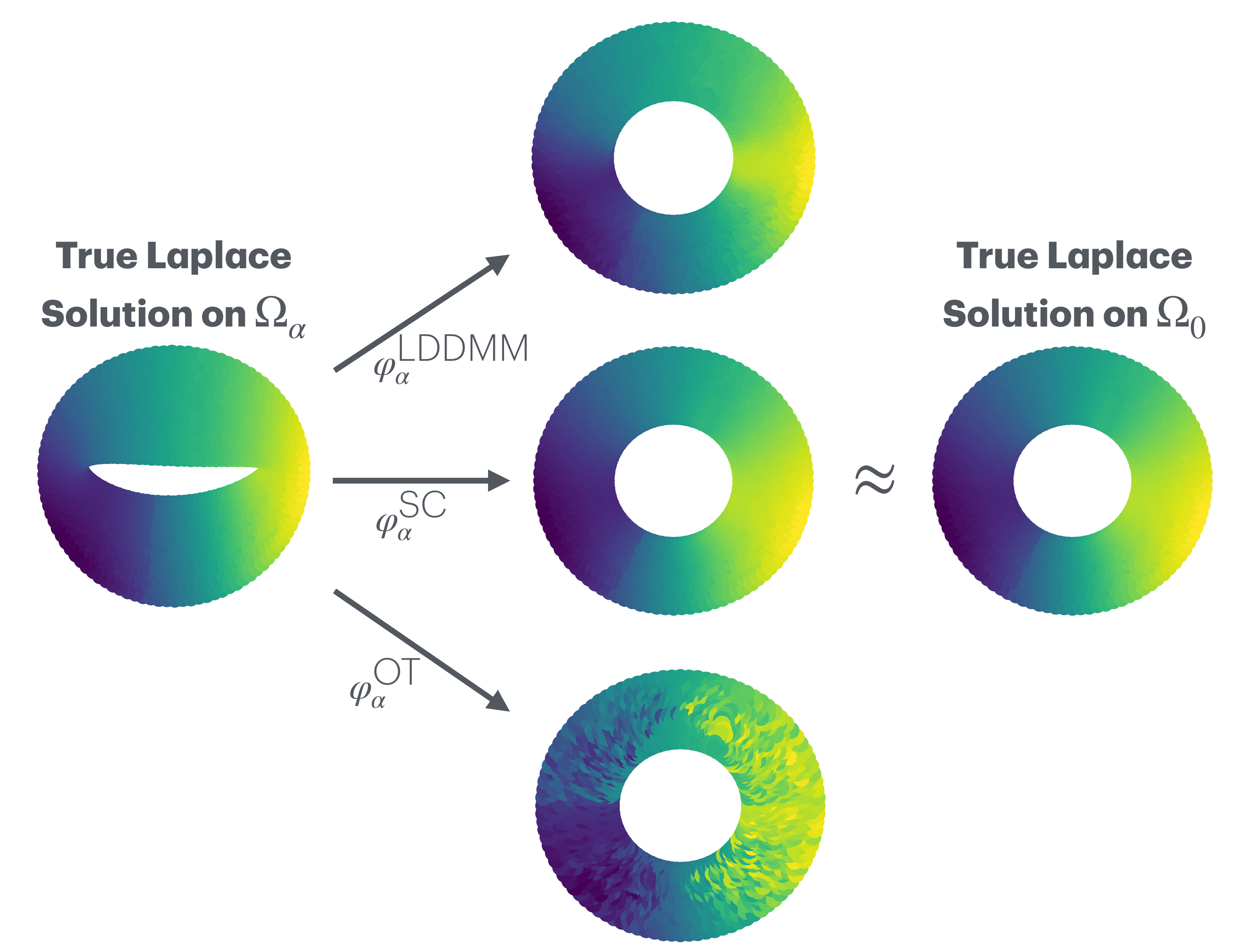}
    \caption{Visual comparison of mappings $\varphi_\alpha : \Omega_\alpha \to \Omega_0$ for different approaches—LDDMM, Schwarz-Christoffel (SC) map, and discrete optimal transport (OT)—and their impact on the Laplace solution. The true Laplace solution on the original domain $\Omega_\alpha$ is mapped to the reference domain $\Omega_0$ using each mapping method. The SC map preserves the Laplace operator, resulting in an exact solution on $\Omega_0$, while LDDMM introduces minor distortions, and discrete OT introduces significant artifacts due to its non-diffeomorphic nature. The rightmost solution represents the true Laplace solution on the annulus $\Omega_0$, illustrating the fidelity of each mapping.}
    \label{fig:mapped-solutions}
\end{figure}

In this section, we evaluate the effectiveness of the three different mapping approaches—conformal maps, Large Deformation Diffeomorphic Metric Mapping (LDDMM), and discrete optimal transport (OT)—for constructing the mappings \(\varphi_\alpha : \Omega_\alpha \rightarrow \Omega_0\) within the diffeomorphic mapping operator learning framework. The results are summarized in Figures \ref{fig:results-mega} and \ref{fig:mapped-solutions} and Table \ref{tab:results-summary}.

Figure \ref{fig:results-mega} displays the ground truth numerical solutions, network predictions, and errors for three different test cases for the Laplace equation solutions on doubly connected domains. These results highlight how the choice of mapping significantly impacts the prediction accuracy and error distribution.

The mapped solutions for each approach are visualized in Figure \ref{fig:mapped-solutions}, demonstrating how different mappings affect the representation of the solution in the reference domain \(\Omega_0\). Our main observations are summarized below:
\begin{itemize}
    \item Conformal maps preserve the Laplace equation entirely, resulting in mapped solutions on the annulus that match the true Laplace solution with the same original boundary conditions. This property eliminates the dependence on the original domain geometry $\Omega_\alpha$, allowing the neural operator to learn a much simpler mapping.
    \item While diffeomorphic, the LDDMM mapping introduces slight distortions in the mapped solutions, deviating from the true Laplace solution on \(\Omega_0\).
    \item As expected for a non-diffeomorphic mapping, discrete OT introduces significant noise, leading to irregularities in the mapped solutions.
\end{itemize}
These observations confirm that the regularity of the mapped solutions strongly depends on the properties of the mapping. The closer the mapped solutions remain to being true PDE solutions in \(\Omega_0\), the more efficient and accurate the operator training becomes.

Table \ref{tab:results-summary} summarizes the relative \(L_2\) error, training epochs, and the number of PCA modes used for the geometry branch in the neural operator. For conformal maps, no geometry branch was required (\(0\) PCA modes), as the mapping completely factors out the geometry variability and preserves the PDE. Conversely, for LDDMM and discrete OT, we used 10 PCA modes of the displacement vector field between \(\Omega_\alpha\) and \(\Omega_0\).

For training, the operator network utilized a branch network for the boundary conditions of the outer domain and a geometry branch for the original domain encoding (except in the conformal case). The conformal map’s complete preservation of the Laplace operator reduced the problem’s dimensionality, enabling the network to efficiently learn the relationship between the boundary condition and the mapped solution.

Figure \ref{fig:results-mega} (right panel) shows how training sample size affects performance. Even with only 80 training samples, the network trained on conformally mapped solutions achieved relative \(L_2\) errors an order of magnitude lower than LDDMM with 400 samples. As the training samples decreased for LDDMM, performance degradation was more pronounced, likely due to insufficient sampling of geometries to represent the displacement fields effectively.
\begin{table}[ht]
\centering
\[
\begin{array}{|l|c|c|c|}
\hline
\varphi_\alpha:\Omega_\alpha \rightarrow \Omega_0 & \text{PCA Modes} & \text{Epochs} & \text{Rel \(L_2\) Error} \\
\hline
\text{Conformal Map} & 0 & 1,000 & 0.26\% \\
\text{LDDMM} & 10 & 10,000 & 2.56\% \\
\text{Discrete OT} & 10 & 50,000 & 22.4\% \\
\hline
\end{array}
\]
\caption{Comparison of different mapping approaches based on PCA modes, training epochs, and relative $L_2$ error.}
\label{tab:results-summary}
\end{table}

It is clear that the further away the mapped solutions are in \(\Omega_0\) from being solutions of the original PDE problem that was solved on \(\Omega_\alpha\) as if it were solved on $\Omega_0$, the less effective training is. Most notably, Figure \ref{fig:results-mega} shows that even with a fifth of the training data (80 samples), the network trained on the conformally mapped solutions is still an order of magnitude better than the results on the 400 training samples mapped to $\Omega_0$ via LDDMM. We also see that as the training samples decrease for LDDMM, the performance degrades at a greater rate, presumably since not enough geometries were sampled to be fully expressive.

In order for this approach to truly factor out the geometric variability in the data, the diagram in Figure \ref{fig:dimon} must commute. Methods which aim to minimize discrepancy from a commuting diagram should achieve better results, faster training times, and require less data to generalize well.
While the differences between the solutions mapped to the reference domain \(\Omega_0\) using LDDMM and conformal maps are subtle, the respective relative \(L_2\) errors are an order of magnitude apart, emphasizing the impact of the mapping's construction on the data efficiency of the neural operator. The LDDMM map introduces minor distortions to the solution field, yet these deviations from the true Laplace solution on \(\Omega_0\) significantly degrade the neural operator's ability to generalize. This finding strongly suggests that even small adjustments to the mapping, particularly in a direction that better preserves the governing differential operator, can yield substantial improvements in the efficiency and accuracy of the latent neural operator \(\mathcal{F}_0\).

Future directions should address the broader challenge of generalizing this approach to more complex and diverse PDE problems. For large-scale, higher-dimensional problems, identifying the optimal mapping \(\varphi_\alpha\) from an uncountably infinite family of potential transformations is highly non-trivial. A promising avenue for exploration lies in the development of parameterized constructions of \(\varphi_\alpha\) that explicitly enforce conservation laws and physical constraints of the solution field when mapping between \(\Omega_\alpha\) and \(\Omega_0\). 

Within the LDDMM framework, one way to achieve this is to modify the velocity field \(\dot{\gamma}\) that drives the mapping to account for the conservation of specific physical quantities, such as mass or energy, during the deformation. This could involve augmenting the optimization problem that determines the velocity field with additional terms encoding these conservation laws. For example, one might enforce pointwise or weak forms of the PDE constraints within the variational formulation used in LDDMM, ensuring that the mapped solution remains closer to the solution of the original PDE. Such an approach would likely improve both the geometric fidelity of the map and the preservation of the physical properties of the solution field.

Additionally, integrating these conservation-driven mappings into the neural operator framework requires careful consideration of how the resulting regularity in \(\Omega_0\) affects the training dynamics and parameterization of the neural network. Understanding the relationship between geometry, physics, and data efficiency is a critical research direction for advancing operator learning methods, especially in data-scarce settings.

\section{Conclusion}

In this work, we numerically demonstrate how the choice of the mapping \(\varphi_\alpha: \Omega_\alpha \to \Omega_0\) impacts the data efficiency of the diffeomorphic latent neural operator \(\mathcal{F}_0\). We evaluate three mapping approaches—conformal maps, LDDMM, and discrete OT—using the Laplace equation as a benchmark on doubly connected domains. The results show that conformal maps, which exactly preserve the differential operator, achieve the highest accuracy and training efficiency with minimal  data samples. In contrast, LDDMM maps introduce slight distortions to the solution on the reference domain, leading to significantly higher relative \(L_2\) errors, while discrete OT maps, being non-diffeomorphic, produce noisy and less effective results.

Our findings highlight that even small deviations in how the mapping preserves the original PDE structure can have dramatic effects on neural operator training and performance. This suggests a strong connection between the geometric properties of the mapping and the latent operator's ability to generalize efficiently. 

Future work will focus on developing (or learning) generalized parameterized constructions for \(\varphi_\alpha\) that incorporate conservation laws and physical constraints. These advances will be crucial for scaling diffeomorphic neural operators to handle diverse, high-dimensional PDEs in practical applications. Ultimately, this study underscores the importance of mapping design as a critical factor in improving the efficiency and accuracy of neural operator frameworks for solving PDEs.

\section{Acknowledgements}
 Z.A. was supported by grant n. 24PRE1196125 - American Heart Association (AHA) predoctoral fellowship and grant n. T32 HL007024 from the National Heart, Lung, and Blood Institute.

\bibliography{aaai25}

\end{document}